\documentclass[sigconf]{acmart}

\usepackage{hyperref}
\usepackage{graphicx} 
\usepackage{subcaption}
\newcommand{\subfigref}[1]{%
  \expandafter\ifx\csname r@#1\endcsname\relax%
    ??\else\ref{#1}\fi
}
\usepackage{booktabs}
\usepackage{multirow}
\usepackage{siunitx}

\usepackage{amssymb}
\usepackage{bbding}
\usepackage{tikz}
\usepackage{adjustbox}
\hypersetup{
    colorlinks=true, 
    urlcolor=magenta,   
}

\AtBeginDocument{%
  }

\setcopyright{acmlicensed}
\copyrightyear{2025}
\acmYear{2025}

\acmConference[Conference acronym 'XX]{Make sure to enter the correct
  conference title from your rights confirmation email}{June 03--05,
  2018}{Woodstock, NY}

\acmConference[CIKM '25] {Proceedings of the 34th ACM International Conference on Information and Knowledge Management}{ November 10--14, 2025}{Seoul, Republic of Korea.}
\acmBooktitle{Proceedings of the 34th ACM International Conference on Information and Knowledge Management (CIKM '25), November 10--14, 2025, Seoul, Republic of Korea}
\acmISBN{979-8-4007-2040-6/2025/11}
\acmDOI{10.1145/3746252.3761241}

\begin{document}

\title{Hierarchy-Consistent Learning and Adaptive Loss Balancing for Hierarchical Multi-Label Classification}

\author{Ruobing Jiang}
\orcid{0000-0003-1209-078X}
\affiliation{%
 \institution{Ocean University of China}
 \city{Qingdao}
 \country{China}}
\email{jrb@ouc.edu.cn}

\author{Mengzhe Liu}
\orcid{0009-0002-7525-397X}
\affiliation{%
  \institution{Ocean University of China}
 \city{Qingdao}
 \country{China}}
\email{liumengzhe@stu.ouc.edu.cn}

\author{Haobing Liu}
\authornotemark[1]
\orcid{0000-0002-2546-3306}
\affiliation{%
  \institution{Ocean University of China}
 \city{Qingdao}
 \country{China}}
\email{haobingliu@ouc.edu.cn}

\author{Yanwei Yu}
\orcid{0000-0002-5924-1410}
\affiliation{%
  \institution{Ocean University of China}
 \city{Qingdao}
 \country{China}}
 \email{yuyanwei@ouc.edu.cn}

\thanks{*Corresponding author.}

\renewcommand{\shortauthors}{Ruobing Jiang, Mengzhe Liu, Haobing Liu and Yanwei Yu}

\begin{abstract}
Hierarchical Multi-Label Classification (HMC) faces critical challenges in maintaining structural consistency and balancing loss weighting in Multi-Task Learning (MTL). 
In order to address these issues, we propose a classifier called HCAL based on MTL integrated with prototype contrastive learning and adaptive task-weighting mechanisms. The most significant advantage of our classifier is semantic consistency including both prototype with explicitly modeling label and feature aggregation from child classes to parent classes. 
The other important advantage is an adaptive loss-weighting mechanism that dynamically allocates optimization resources by monitoring task-specific convergence rates. It effectively resolves the "one-strong-many-weak" optimization bias inherent in traditional MTL approaches. 
To further enhance robustness, a prototype perturbation mechanism is formulated by injecting controlled noise into prototype to expand decision boundaries. 
Additionally, we formalize a quantitative metric called  Hierarchical Violation Rate (HVR) as to evaluate hierarchical consistency and generalization. Extensive experiments across three datasets demonstrate both the higher classification accuracy and reduced hierarchical violation rate of the proposed classifier over baseline models.
\end{abstract}




\maketitle

\begin{figure}
    \centering
    \includegraphics[width=1\linewidth]{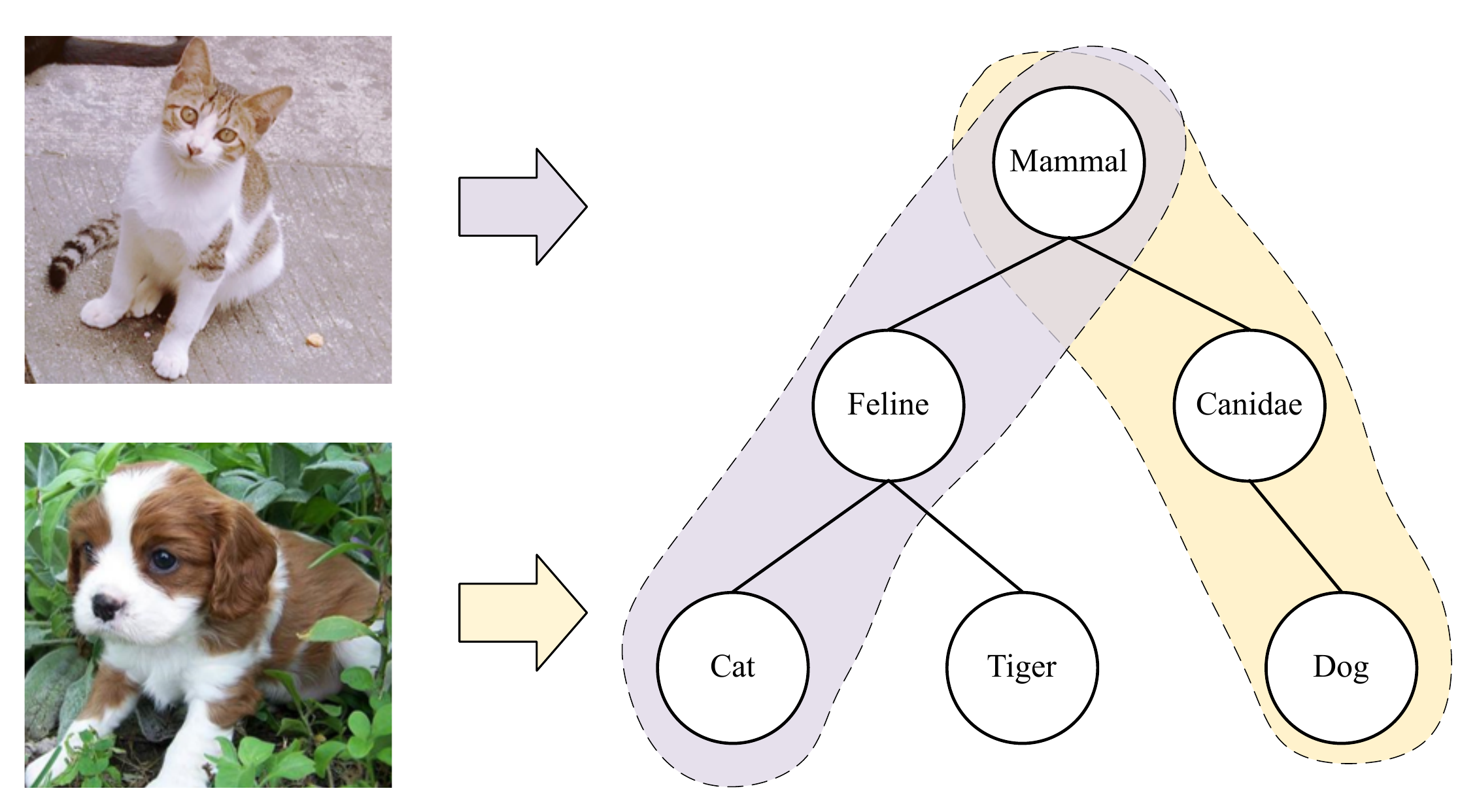}
    \caption{An example with tree-structured parent-child label relationships. The left panel contains concrete instances: a cat (upper) and a dog (lower) as terminal node exemplars while the right tree structure comprises six hierarchical nodes connected by parent-child relationships. Instance-to-label mappings are annotated with same color arrows and boxes.}
    \label{fig:1}
\end{figure}

\section{Introduction}
Hierarchical Multi-Label Classification (HMC) ~\cite{CHMCN-Nips2020,f-EDR-CIKM2024} addresses the inherent hierarchical dependencies in real-world labeling systems, where labels exhibit tree-structured parent-child relationships. As shown in Figure \ref{fig:1}, an image may contain multiple hierarchical labels such as "cat", "feline", and "mammal". These dependencies enforce strict structural constraints: predictions must preserve semantic consistency across hierarchy levels ~\cite{C-HMCNN-Journal} (e.g., classifying an image as "cat" necessitates correct higher-level assignments to "feline" and "mammal").

Existing HMC methods \cite{CHMCN-Nips2020,C-HMCNN-Journal,HCMN} mainly adopt two distinct frameworks. Some approaches treat hierarchical classification as a single-task problem, integrating all hierarchical levels prediction into a unified model ~\cite{HmcNet-2023TKDE}. For example, HyperIM ~\cite{hyperim} embeds text representations and labels representations in hyperbolic space to leverage geometric properties for hierarchy preservation, where the exponential growth of distances between nodes in hyperbolic geometry naturally preserves hierarchical relationships. However, single-task methods exhibit inherent limitations in capturing intricate interdependencies across hierarchical levels. In contrast, Multi-Task Learning (MTL) ~\cite{MTL1997, MTL-TKDE2022} framework devides HMC into hierarchical subtasks, where each level trains its specialized classifier. These MTL-based methods ~\cite{MTLRecommand-CIKM2024,MTLora-CVPR2024} usually enforce parent-child relationships through shared parameters.

However, MTL-based methods face two core challenges that interact synergistically in HMC: \textit{how to ensure consistency of label prediction between parent and child classes} and \textit{how to balance task-weighting dynamically}. First, existing MTL-based methods treat each label as an independent task, ignoring hierarchical dependencies, which leads to high parent-child prediction conflict rates. For instance, on the CIFAR-100 dataset, MMoE ~\cite{MMOE-KDD2018} exhibits a hierarchical violation rate (i.e., inconsistencies between child and parent class predictions) as high as 10\%, while DMT ~\cite{DMT-CIKM2020} also demonstrates a substantial violation rate of nearly 12\%. Second, due to the different loss scales or gradient magnitudes of various tasks ~\cite{IMTL-ICLR2021, ConflictAverse-nips2021a, Nash-ICML2022}, MTL-based methods creat an imbalanced "one-strong-many-weak" state. Although a common strategy is to manually assign task-specific loss weights to mitigate this issue, this approach suffers from clear limitations in practice. Manually tuning these weights is labor-intensive, heavily relies on exhaustive hyperparameter experiments, and struggles to adapt to dynamically changing training processes.

To address the first challenge, we propose a MTL-based classifier called HCAL with prototype contrastive learning for hierarchical semantic consistency modeling. Departing from conventional contrastive methods ~\cite{CL-CCFC2024} that optimize individual instances in isolation, this method constructs learnable prototype representations in the feature space to explicitly characterize the semantic properties of all classes of all levels. Meanwhile, parent-class features are dynamically updated by aggregating features from their child-class counterparts, forming an interpretable semantic hierarchical dependencies. This design enables the model to jointly low and high levels categorical relationships, thereby ensuring logical consistency in hierarchical predictions at the representation learning level. To further enhance classification robustness, we design a prototype boundary perturbation mechanism that injects controlled noise into prototype vectors. It can adaptively expand classification decision boundaries in the feature space to mitigate misclassifications caused by data distribution shifts or adversarial perturbations. 

For the second challenge, we devise an adaptive loss weight adjustment mechanism in the training process to achieve dynamic task-balancing optimization. This mechanism monitors the convergence rates of learning curves across different tasks and dynamically adjusts their loss weights, preferentially allocating optimization resources to underfitting tasks. This approach overcomes the limitations of static weighting strategies. Additionally, we innovatively introduce a quantitative evaluation metric called Hierarchical Violation Rate (HVR)  to measure the adherence of predictions to hierarchical labels structures.

In summary, our contributions are as follows.

\begin{itemize}
\item {\texttt{}}
We formulate a prototype-based contrastive learning classifier called HCAL for HMC, where category-specific prototypes explicitly model labels. Meanwhile, parent-class features are come from child-class features, forming an interpretable and consistent hierarchy in the representation space.
\item {\texttt{}}
We design an adaptive loss weight adjusting mechanism that dynamically balances task contributions by aligning gradient optimization and synchronizing convergence rates. By monitoring the optimization trajectories of different levels tasks, this strategy adaptively adjusts loss weights.
\item {\texttt{}}
We develop a prototype perturbation method with controlled noise injection. Through perturbed prototype contrastive learning, the model gains enhanced robustness.
\item {\texttt{}}
We propose a novol metric called HVR to quantify hierarchical structural consistency and validate the method’s efficacy. Experiments across multi-domain datasets demonstrate that our approach achieves state-of-the-art performance in both classification accuracy and hierarchical constraint adherence.
\end{itemize}

\section{Related Work}
\subsection{Hierarchical Multi-Label Classification}
In HMC, existing approaches primarily include single-task and multi-task learning.

Single-task methods unify hierarchical classification into a single framework, directly modeling label dependencies. For instance, HyperIM ~\cite{hyperim} embeds text and labels in hyperbolic space to optimize predictions through geodesic distances while HRN ~\cite{HRN} designs a hierarchical residual network using residual connections to propagate parent-child features. The Hierarchy-aware Label Semantics Matching Network ~\cite{HLSMN} aligns text representations with hierarchical label semantics through a matching network, strengthening parent-child semantic inheritance. These methods reduce complexity reduces complexity but struggles with dynamic hierarchies. Conversely, MTL explicitly partitions tasks across hierarchical levels and treats each level of the hierarchy as an independent task, where dedicated models are employed for prediction at each level. Representative works include HMCN ~\cite{HCMN}, which balances local classifiers and global objectives. C-HMCN further introduces the MCLoss to post-processing the results. MHCAN ~\cite{MHCAN} leverages cross-level attention to capture label dependencies. HSVLT ~\cite{hsvit} develops a hierarchical scale-aware vision-language Transformer for multi-label image classification, integrating hierarchical priors into cross-modal alignment. They aim to mitigate prediction inconsistencies, with MTL emphasizing explicit task coordination and single-task methods prioritizing architectural simplicity.

\subsection{Multi-Task Learning}
MTL aims to optimize multiple tasks simultaneously through shared parameters ~\cite{MTLRecommand-CIKM2024}. 

Researchers firstly focus on structural modifications to balance shared and task-specific characteristics among tasks. A typical multi-task architecture design comprises an encoder that extracts feature representations from input frames and a set of task-specific decoders generating predictions for each task ~\cite{MTL-TKDE2022}. The main concern in the MTL architecture is how to share information with other tasks ~\cite{MTLora-CVPR2024,TKDD2022}. The hard parameter sharing ~\cite{MTL1997} uses a shared backbone network to enable the sharing of knowledge between tasks, while each task maintains independent decoders to generate its predictions ~\cite{ConflictAverse-nips2021a}. MT-DNN ~\cite{HardParaShare-ACL2019fromSoftware} exemplifies this paradigm. But hard parameter sharing struggles to effectively leverage cross-task information, particularly the hierarchical relationships between different labels. Soft parameter sharing implements independent models for distinct tasks while establishing inter-task parameter constraints through mechanisms. Multi-gate Mixture-of Experts (MMoE) ~\cite{MMOE-KDD2018} replaces the single shared layer with multiple expert networks, dynamically combining their outputs through a gating network that assigns input-dependent weights to experts. DMT ~\cite{DMT-CIKM2020} employs multiple Transformers to simultaneously model diverse user behavior sequences and also leverages the MMoE framework to optimize multiple objectives. DAPAMT \cite{TKDD2022} designs co-attention-based task interaction units to explicitly transfer knowledge across academic prediction tasks, addressing data sparsity and task correlation challenges.

Another approach optimizes MTL by resolving gradient conflicts in shared parameters. When gradients align directionally, they enhance learning efficiency. GradNorm ~\cite{gradnorm-ICML2018} balances the magnitudes of gradients in tasks, and PCGrad ~\cite{PCGrad-nips2020} resolves conflicting gradients through geometric projection. Nash-MTL ~\cite{Nash-ICML2022} formulates MTL as a cooperative bargaining game under game-theoretic principles.

\subsection{Prototype Learning}
Prototype learning is a machine learning paradigm that operates by maintaining a set of representative prototype vectors to perform classification, regression, or clustering tasks. This methodology mimics human cognitive processes in concept formation and has been explored in a variety of tasks. It can be used to robustly handle complex data distributions by explicitly modeling category centroids and decision boundaries in the embedding space, and it has been explored in a variety of tasks ~\cite{FedProto-AAAI2022}. 

In continual learning ~\cite{OnlinePro}, ProtoSRD ~\cite{ProtoSRD} employs prototypes to regulate inter-sample distance discrepancies in the feature space, mitigating catastrophic forgetting in neural networks. For federated learning, FedProto ~\cite{ferProto} aligns heterogeneous data via weighted aggregation of cross-client category prototypes to reduce communication costs. In few-shot learning ~\cite{fewshotProto}, PrototypeFormer ~\cite{ProtoTransformer2025} enhances the separability between classes by integrating Transformer with prototype learning, reaching a high precision. And for unsupervised learning, SIGMA ~\cite{SIGMA} introduces a novel graph-matching-based prototype alignment framework for cross-domain adaptation, and achieves some improvement on the domain adaptation benchmark.

\subsection{Contrastive Learning}
Contrastive learning, a self-supervised approach, learns representations by contrasting positive and negative sample pairs in feature space, demonstrating strong generalization across domains such as computer vision and NLP ~\cite{CL-CCFC2024}.

Its core objective is to minimize the distance between positive pairs while maximizing it for negative pairs, which is called Info Noise-Contrastive Estimation Loss (InfoNCE) ~\cite{infonce}. In computer vision, SimCLR achieves accuracy on ImageNet through multi-view augmentation and nonlinear projection heads ~\cite{simclr}. For NLP, SimCSE generates positive pairs via stochastic dropout masks, significantly enhancing text embedding quality on STS benchmarks ~\cite{simcse}. A-InfoNCE ~\cite{Ainfonce} introduces an asymmetric contrastive loss that treats adversarial examples as "weak positives" or "hard negatives" to address the conflict between adversarial training and contrastive learning objectives.

\begin{figure*}
    \centering
    \includegraphics[width=1\textwidth]{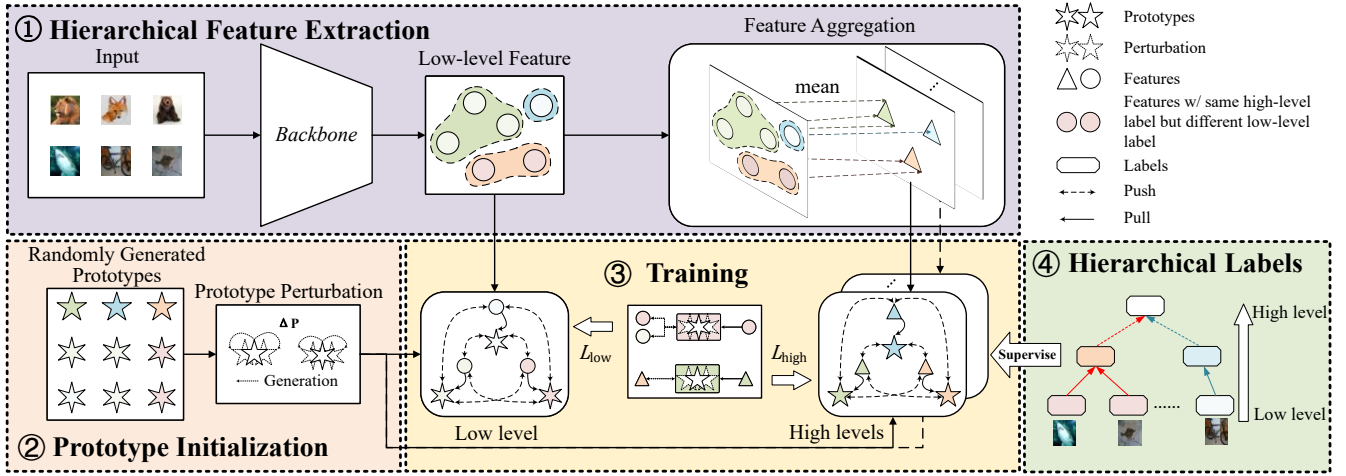}
    \caption{Overview of our approach. It includes three modules: hierarchical feature extraction, prototype initialization and training. The rounded rectangles represent the current state, while rectangular shapes with sharp corners denote the current operations. The icons of different colors or shaping denote the different classes.}
    \label{fig:framework}
\end{figure*}

\section{The Proposed Method}
We describe the proposed approach called HCAL in Figure~\ref{fig:framework}. The proposed method operates through four stages. In hierarchical feature extraction module, input samples are processed through the backbone network to generate multi-level feature representations. In the stage of prototype initialization, category-specific prototype vectors are initialized in the embedding space. Then in training, features and prototypes are jointly optimized using a hierarchical contrastive loss, where hierarchical labels are used to supervised. Finally, outputs are hierarchically structured predictions.
\subsection{Preliminaries}
Given a training set $X=\{x_1,x_2,...,x_n \}$ of $n$ images, whose labels are $Y=\{Y_1,Y_2,...,Y_n\}$. The sample $x_i$ has multiple labels as $Y_i=\{y^1_i,y^2_i,...,y^m_i\}$ with $i\in \{1,2,...n\}$ and $m$ denoting the total number of hierarchical levels. For each hierarchical level $k\in\{1,2,...,m\}$, we define the set of labels at level $k$ as $\mathcal{C}^{k}=\bigcup_{i=1}^{n}\left\{y_{i}^{k}\right\}$. For each label $c\in \mathcal{C}^{k}$, we define its corresponding set of samples as $S_c^k=\{x_i\in X | y^k_i=c\}$, where $S_c^k$ denotes the set of all samples assigned to label $c$ at the hierarchical level $k$. Let $|\mathcal{C}^{k}|$ denotes the total number of labels at the $k$-th hierarchical level, and $|Y|=\sum_{k=1}^{m} |\mathcal{C}_{k}|$ represents the aggregate count of labels across all $m$ hierarchical levels. Similarly $|S_c^k|$ is also the total number of samples annotated with $c$ at the level $k$.

\subsection{Hierarchical Feature Extraction}
In the first stage of our model, the primary objective is to generate hierarchical aligned feature representations for all samples across multiple label levels $F=\{F^k|k\in \{1,2,...,m\}\}$. We adopt ResNet18 as the backbone network for the preliminary extraction of characteristics. 

The input $X$ is fed into the backbone network to extract preliminary feature representations. These low-level features serve as the foundation for subsequent hierarchical contrastive learning. They are represented by $F^1=\{f^1_i|i\in \{1,2,...,n\}\} \in \mathbb{R}^{n \times d}$, where $d$ denotes feature dimension. 
Then, we perform hierarchical feature aggregation to construct high-level representations $\{F^2,F^3,...,F^m\}$ corresponding to $m$ label hierarchy levels.  Inspired by tree-structured operations, each level $k>1$ aggregates features from the previous level $F^{k-1}$, that is, $F^k={\rm Aggregation}(F^{k-1})$.

Specifically, we partition the feature representations $F^{k-1}$ of all samples at the hierarchical level $k-1$ into disjoint subsets based on the consistency of the label at the level $k$. Formally,
\begin{equation}
    F^{k-1}=\bigcup_{c\in \mathcal{C}^{k}}{F^{k-1}_c},
\end{equation}
where $F_{c}^{k-1}=\left\{f_{i}^{k-1} \mid y_{i}^{k}=c\right\}$, and $F_{c}^{k-1}$ denotes the subset of features from samples sharing the same label $c$ at level $k$. Next, we aggregate features sharing the same parent-class label by computing their mean as the parent-class representation. The feature representation $F^k_{c'} \in \mathbb{R}^{\mathcal{C}^{k} \times d}$ for parent class $c'$ at level $k$ is derived by averaging all corresponding child-class features from level $k-1$:
\begin{equation}
    F^k_{c'}=\frac{1}{|S_{c'}^{k-1}|}\sum_{c\in {\mathcal{C}_{c'}^{k-1}}}F_{c}^{k-1},
    \label{aggregation}
\end{equation}
where $\mathcal{C}^{k-1}_{c'}$ denotes the set of child classes under parent class $c'$. When $k=m$, feature aggregation is terminated.

\subsection{Prototype Initialization}
The prototype initialization process comprises two interdependent phases, randomly generating prototypes and perturbation. 

First, a stochastic sampling strategy is used to generate the original prototype vectors as $\mathbf{P}=\{\mathbf{p}_j\}\in \mathbb{R}^{|Y| \times d}$, where $\mathbf{p}_j$ indicates the prototype for the $j$-category. 
To enhance the robustness of decision boundaries in the feature space, we introduce a uniformly distributed perturbation mechanism. For each prototype $\mathbf{p}_j$, the perturbation is defined as:
\begin{equation}
    \tilde{\mathbf{p}_j}=\mathbf{p}_j+\Delta \mathbf{P}_j, \Delta \mathbf{P}_j \sim \mathcal{U}(-\epsilon ,\epsilon )^d ,
\end{equation}
where $\epsilon$ denote the perturbation strength hyperparameter, governing the sampling boundaries of the injected noise. The perturbed prototypes are $L_2$-normalization to maintain unit hypersphere constraints:
\begin{equation}
    \tilde{\mathbf{p}}_{j} \leftarrow \frac{\tilde{\mathbf{p}}_{j}}{\left\|\tilde{\mathbf{p}}_{j}\right\|_{2}},
\end{equation}
This operation introduces controllable randomness in tangential space while preserving the original semantic direction, which is equivalent to performing data enhancement.

\subsection{Training with Adaptive Loss Weight Adjustment}
In this stage, we integrate the hierarchical prototypes and feature representations obtained from the preceding stages. The training process jointly optimizes backbone and prototypes. For backbone, we adjust the pre-trained backbone by fine-tuning and train its final fully connected layer while dynamically updating prototypes to align with evolving feature distributions. 

Our training method is adopting contrastive learning between features and prototypes at the same level with supervising by hierarchical labels. For each sample $x_i$, its positive sample set is prototype $\mathcal{P}_i^k=\{\mathbf{p}_j^k\} \bigcup  \{\tilde{\mathbf{p}}_{j}\}$ at the $k$-level, where $\mathbf{p}_j^k$ is the original prototype and $\tilde{\mathbf{p}}_{j}$ is its perturbation for the class $j$, and the negative sample set are all prototypes at the level $k$ excluding the class $j$, which is $\mathcal{N}_i^j$. We calculate the distance between positive and negative sample pairs by using cosine similarity:
\begin{equation}
    {\rm sim}(x_i,\mathbf{p})=\frac{f_i^\top \cdot \mathbf{p}}{||f_i||_2 \cdot ||\mathbf{p}||_2} ,
    \label{con:cosin}
\end{equation}
where $f_i$ is the feature representation of $x_i$. When $k=1$, we maximize the mutual information between samples and corresponding prototypes through InfoNCE loss:
\begin{equation}
    \mathcal{L}_{1}^{}=-\frac{1}{n} \sum_{\mathbf{p} \in \mathcal{P}_{i}^{1}} \log \frac{\exp \left(\operatorname{sim}\left(x_{i}, \mathbf{p}\right) / \tau\right)}{\sum_{\mathbf{q} \in \mathcal{P}_{i}^{1} \cup \mathcal{N}_{i}^{1}} \exp \left(\operatorname{sim}\left(x_{i}, \mathbf{q}\right) / \tau\right)},
    \label{con:loss-1}
\end{equation}
where $\tau$ is temperature coefficient to controls the sharpness of similarity distributions. Similarly, the calculation of loss function of the level of $k>1$ by:
\begin{equation}
    \mathcal{L}_{k}^{}=-\frac{1}{|\mathcal{C}^k|} \sum_{\mathbf{p} \in \mathcal{P}_{i}^{k}} \log \frac{\exp \left(\operatorname{sim}\left(x_{i}, \mathbf{p}\right) / \tau\right)}{\sum_{\mathbf{q} \in \mathcal{P}_{i}^{k} \cup \mathcal{N}_{i}^{k}} \exp \left(\operatorname{sim}\left(x_{i}, \mathbf{q}\right) / \tau\right)},
    \label{con:loss-k}
\end{equation}
which is optimized by two coupling mechanisms: the alignment of features and prototypes and the robustness of prototypes-disturbed prototypes. Finally, our total loss is a combination of the losses introduced above:
\begin{equation}
    \mathcal{L}_{total}=\sum_{k=1}^{m} \lambda_k \cdot \mathcal{L}_k ,
\end{equation}
where $\lambda_k$ denotes the loss weight of level of $k$.

\begin{figure}
    \centering
    \includegraphics[width=0.9\linewidth]{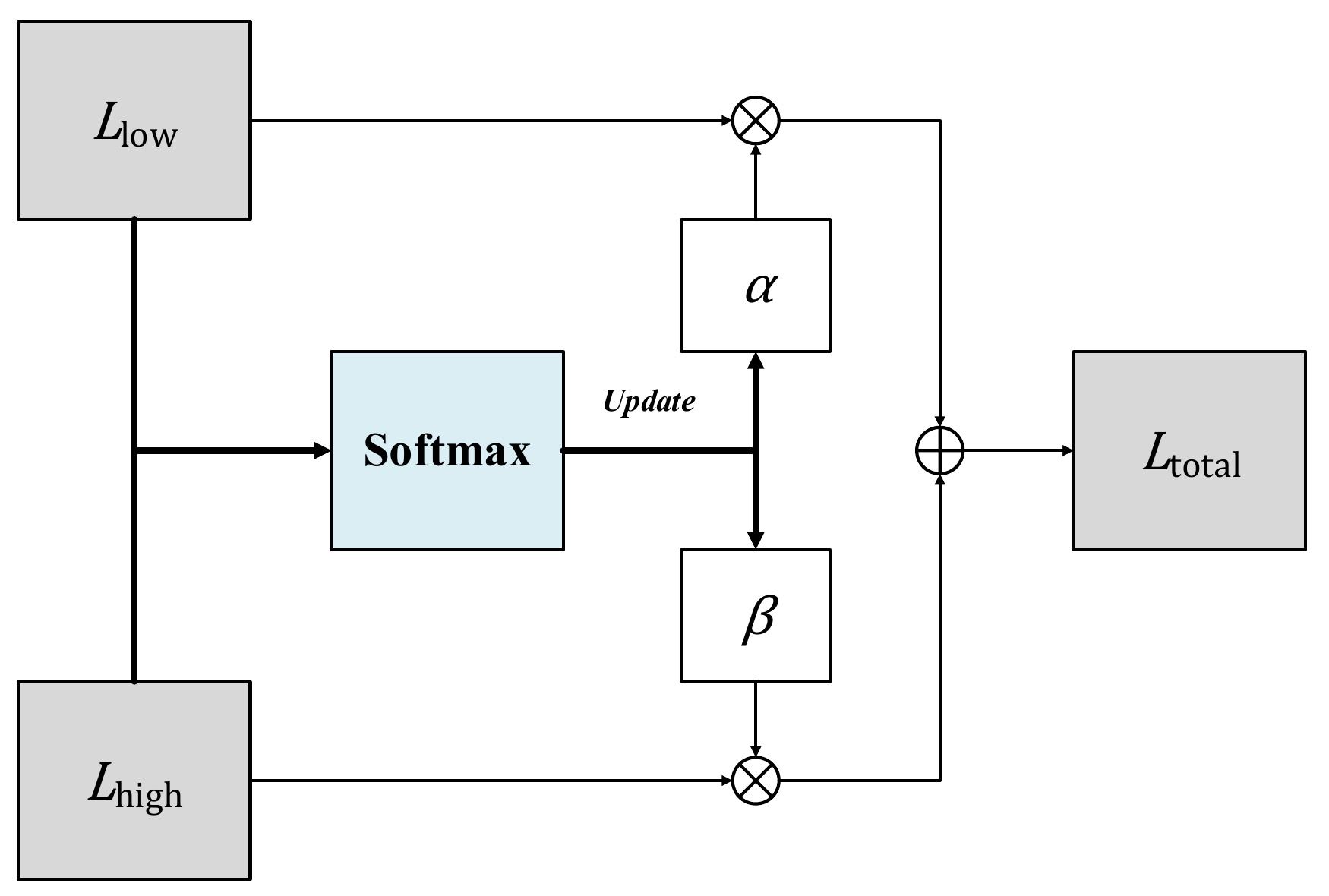}
    \caption{Adaptive weight adjustment mechanism when $m=2$. $L_{\rm low}$ at upper-left and $L_{\rm high}$ at lower-left fed in parallel, converge to central Softmax operator for weight normalization. Parameters $\alpha$ and $\beta$ are updated by softmax outputs, and then are calculated with losses to produce total loss $L_{\rm total}$.}
    \label{fig:loss}
\end{figure}

However, manually tuning the weight parameters of the loss functions proves to be a labor intensive and suboptimal practice. To address this challenge, we propose an adaptive weight adjustment mechanism that automatically balances multi-level learning objectives. For methodological clarity, we assume the total number of hierarchical levels $m=2$, where the lower-level loss $\mathcal{L}_{\rm low}$ and the higher-level loss $\mathcal{L}_{\rm high}$ are computed as specified in Eq. ~\eqref{con:loss-1} and Eq. ~\eqref{con:loss-k}.  The total optimization objective is formulated through weighted fusion of hierarchical losses:
\begin{equation}
    \mathcal{L}_{\rm total}=\alpha\cdot\mathcal{L}_{\rm low}+\beta\cdot{L}_{\rm high},
\end{equation}
where $\alpha,\beta \in [0,1]$ denotes adaptive weights satisfying $\alpha+\beta=1$. The workflow of this adaptive loss fusion and parameter update mechanism is illustrated in Figure \ref{fig:loss}. 

The core innovation lies in dynamically adjusting the weight coefficients $\alpha$ and $\beta$ to coordinate the optimization priorities between hierarchical objectives, driving the refinement of the iterative parameters. This process establishes a closed-loop feedback system: after each parameter update, the forward propagation guide automatic recalibration of two weight parameters. Specially, our proposed adaptive weight adjustment operates through the following iterative procedure. First, get the losses from previous processes. Second, transform loss values into normalized weights via softmax function:
\begin{equation}
    \alpha= \frac{{\rm exp}(\mathcal{L}_{\rm low}/\gamma )}{{\rm exp}({\mathcal{L}_{\rm low}/\gamma})+{\rm exp}({\mathcal{L}_{\rm high}/\gamma})},
\end{equation}
where $\gamma$ is a hyperparameter governing the smoothness of softmax function. Let $m>2$, we can calculate weights by:
\begin{equation}
    \lambda_i=\frac{{\rm exp}(\mathcal{L}_i/\gamma)}{\sum^{m}_{k=1}{\rm exp}(\mathcal{L}_k/\gamma)}.
    \label{con:loss-all}
\end{equation}

\subsection{Prediction} 
Eventually, during inference, the test sample set is processed through the hierarchical feature extraction module to obtain multi-level feature representations. In particular, the test samples are first processed through the backbone network to extract feature representations corresponding to the lowest-level prototypes. Subsequently, the higher-level features are computed from the bottom up by recursively applying Eq. ~\eqref{aggregation} to aggregate lower-level features across layers.

For each hierarchical level, the prediction is generated by comparing the extracted features with the learned prototypes via Eq. ~\eqref{con:cosin}, followed by nearest-prototype classification. The model will generate a hierarchically structured prediction for each sample:
\begin{equation}
    \hat{\mathbf{y}}_{i}=\left(\hat{y}_{i}^{1}, \hat{y}_{i}^{2}, \ldots, \hat{y}_{i}^{m}\right),
\end{equation}
where $\hat{y}_{i}^{k}$ represents the predicted label at the $k$-th level.

\section{Experiment}
This section is organized as follows: Section 4.1 details the foundational components of our experiments, including dataset descriptions, evaluation protocols, and implementation configurations. Section 4.2 conducts a comparative analysis of classification performance against state-of-the-art baselines. Section 4.3 validates the contribution of individual core modules through systematic ablation studies. Section 4.4 examines the impact of two critical hyperparameters. Finally, Section 4.5 presents visualization of feature-prototype distribution.

\begin{figure*}
\centering
\includegraphics[width=1\linewidth]{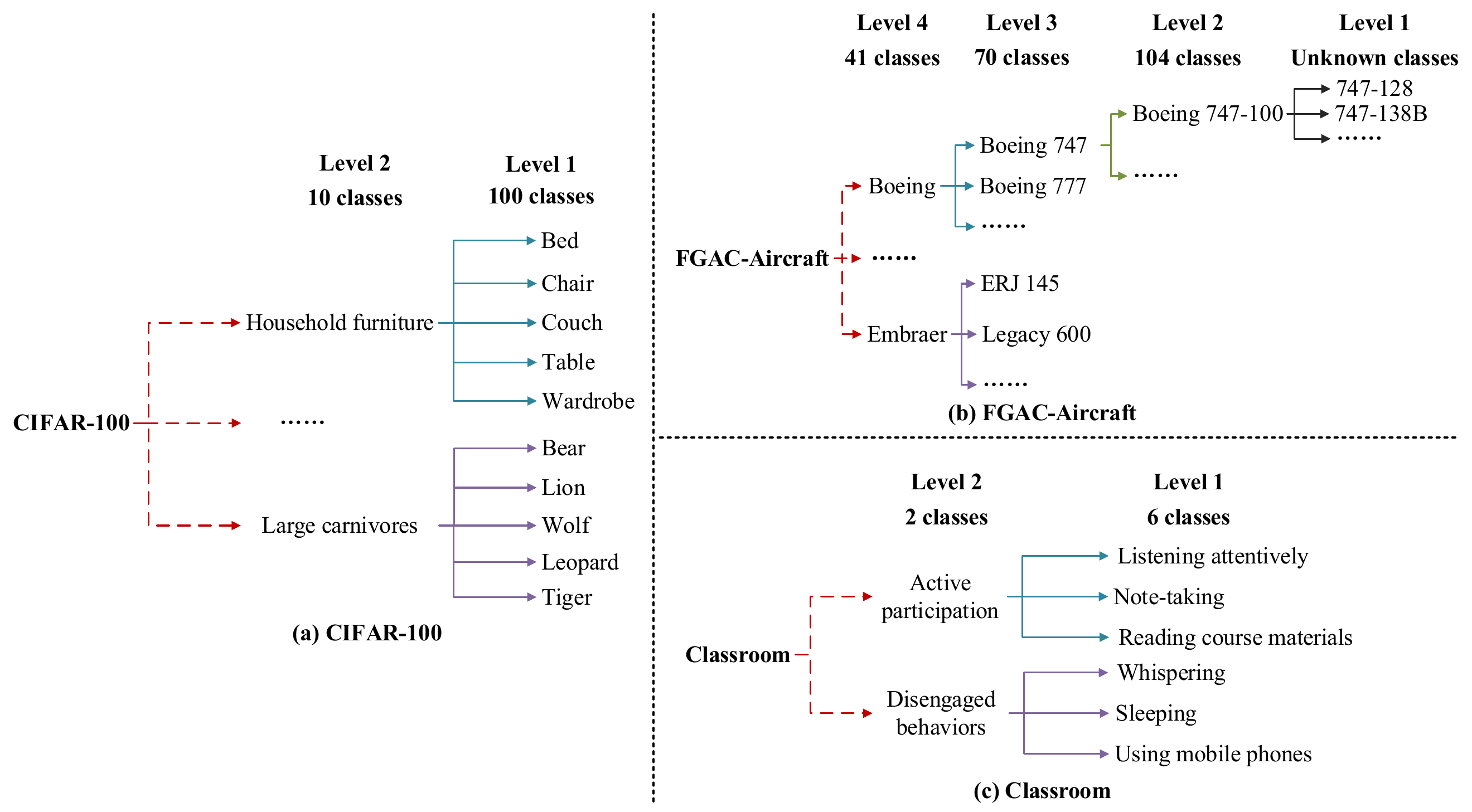}
\caption{The tree-structure of three datasets. We encode the hierarchical labels of the datasets in a bottom-up manner (represented from right to left in the figure). Both the CIFAR-100 and Classroom datasets contain two hierarchical levels. In contrast, the FGVC-Aircraft dataset comprises four hierarchical levels, whose first level lacks its category count.}
\label{fig:tree}
\end{figure*}

\subsection{Experimental Settings}
\subsubsection{Dataset}

The experiments employ three datasets, including CIFAR-100 \cite{cifar100dataset}, FGVC-Aircraft \cite{aircraft} and Classroom. Their tree-structures are shown in Figure \ref{fig:tree}.

{\bf CIFAR-100}. Each image in this dataset is annotated with two hierarchical labels: a fine-grained label and a coarse-grained label. The dataset contains 100 fine-grained classes, with each class comprising 600 images (500 for training and 100 for testing). The coarse-grained labels act as superclasses, grouping the 100 fine-grained classes into 20 broader categories. This hierarchical structure enables multi-level classification, where fine-grained classes (e.g., cat) are nested under their corresponding coarse-grained superclasses (e.g., animals). Part of the specific label structure is shown in the Figure \ref{fig:tree}(a).

{\bf FGVC-Aircraft}. This benchmark dataset also called Fine-Grained Visual Classification of Aircraft, is designed for fine-grained visual categorization of aircraft models. It contains 10,200 images covering 102 distinct aircraft variants, with 100 images per variant. Each image is annotated with a tight bounding box and hierarchical labels organized into four levels: model, variant, family, and manufacturer. The dataset is equally partitioned into three subsets for training, validation, and testing, ensuring balanced evaluation across all aircraft categories. 
In Figure \ref{fig:tree}(b), we give four levels of partial labels. In the subsequent experiments, we didn't use level 1 to compare with others.

{\bf Classroom}. The Classroom dataset is collected from a university, with all images undergoing privacy-preserving processing. This custom-built dataset captures student engagement states in authentic classroom settings, comprising 20,500 high-resolution images (224×224 pixels). It is partitioned into 20,000 training images and 500 test images. We establish a two-tier hierarchical taxonomy. The primary level has two categories, active participation and disengaged behaviors. In the secondary level, the former includes listening attentively, note-taking, reading course materials to interact while the latter has whispering, sleeping and using mobile phones three classes. 
The tree-structure labels of the dataset is shown in the Figure \ref{fig:tree}(c).

\begin{table*}[htbp]
    \centering
    \caption{Comparison with other methods on three databases. The upward arrow indicates that the larger the data, the better the performance of the model, while the downward arrow indicates that the smaller the data, the better the model. Bold data represents the best.}
    \label{table:compare}
    \begin{tabular}{@{}l *{10}{S[table-format=1.4]}@{}}
    \toprule
    \multirow{2}{*}{\textbf{Method}} & 
    \multicolumn{3}{c}{\textbf{CIFAR100}} & 
    \multicolumn{4}{c}{\textbf{FGVC-Aircraft}} & 
    \multicolumn{3}{c}{\textbf{Classroom}} \\
    \cmidrule(lr){2-4} \cmidrule(lr){5-8} \cmidrule(l){9-11}
    & \multicolumn{1}{c}{$ACC^1$↑} 
    & \multicolumn{1}{c}{$ACC^2$↑} 
    & \multicolumn{1}{c}{$HVR$↓} 
    & \multicolumn{1}{c}{$ACC^2$↑} 
    & \multicolumn{1}{c}{$ACC^3$↑}
    & \multicolumn{1}{c}{$ACC^4$↑}
    & \multicolumn{1}{c}{$HVR$↓} 
    & \multicolumn{1}{c}{$ACC^1$↑} 
    & \multicolumn{1}{c}{$ACC^2$↑} 
    & \multicolumn{1}{c}{$HVR$↓} \\
    \midrule
    MMoE    & 0.5181 & 0.6594 & 0.0975 & 0.5521 & 0.7402 & 0.8497 & 0.1602 & 0.9301 & 0.9696 & 0.0160 \\
    DMT     & 0.5905 & 0.7150 & 0.1149 & 0.4821 & 0.6544 & 0.7732 & 0.2772 & 0.9251 & 0.9610 & 0.0171 \\
    CAGrad  & 0.5850 & 0.7233 & 0.1059 & 0.6328 & 0.7864 & 0.8695 & 0.1116 & 0.9229 & 0.9544 & 0.0188 \\
    IMTL    & 0.5908 & 0.7208 & 0.1132 & 0.6361 & 0.7765 & 0.8755 & 0.1176 & 0.9340 & 0.9705 & 0.0163 \\
    Nash-MTL& 0.5940 & 0.7219 & 0.1098 & 0.6262 & 0.7921 & 0.8707 & 0.1137 & 0.9332 & 0.9761 & 0.0179 \\
    CMT & 0.7340 & 0.8426 & 0.0601 & 0.5581 & 0.7486 & 0.8683 & 0.1299 & 0.9458 & 0.9801 & 0.0135 \\
    HCAL    & \textbf{0.7480} & \textbf{0.8458} & \textbf{0.0582} & \textbf{0.6829} & \textbf{0.8233} & \textbf{0.8806} & \textbf{0.1012} & \textbf{0.9491} & \textbf{0.9807} & \textbf{0.0114} \\
    \bottomrule
    \end{tabular}
\end{table*}

\subsubsection{Evaluation Metrics}
 Accuracy is a fundamental evaluation metric in machine learning and statistics, measuring the proportion of correctly classified instances relative to the total samples. In our experiments, the accuracy at the $k$-level is defined as $ACC^k$:
\begin{equation}
    ACC^k=\frac{1}{n} \sum_{i=1}^{n} \mathbb{I}(y_i^k=\hat{y}_i^k),
\end{equation}
where $\mathbb{I}$ is the indicator function, mapping a logical condition to a binary output: returns 1 if the condition within the parentheses holds true and 0 otherwise.

In addition, to quantitatively measure structural consistency in hierarchical multi-label classification, we propose the Hierarchical Violation Rate (HVR) metric. HVR evaluates the proportion of predictions that violate parent-child constraints in the label hierarchy. Formally, for a dataset with $n$ samples and $m$ levels, HVR is defined as:
\begin{equation}
    HVR=\frac{1}{n \cdot R} \sum_{i=1}^{n} \sum_{k=2}^{m} \sum_{c \in \mathcal{C}^{k}} \mathbb{I}\left(\hat{y}_{i}^{k-1}=c \wedge \hat{y}_{i}^{k} \neq p(c)\right),
\end{equation}
where $p(c)$ denotes the parent label of $c$ at level $k$, and $R$ is total parent-child relationships across all levels, computed as:
\begin{equation}
    R=\sum_{k=1}^{m-1}\left|\mathcal{C}^{k}\right|.
\end{equation}

\subsubsection{Implementation Details}
We use ResNet-18 pre-trained on ImageNet as the backbone network to extract hierarchical feature representations. We maintain a consistent batch size of 32 and set the feature dimension $d$ to 256 for all levels and datasets. During training, we employ Stochastic Gradient Descent (SGD) as the optimizer with several different learning rate. The backbone is initialized with a learning rate of 0.01. The lowest level prototypes are optimized with learning rate of 0.05. The prototype learning rate progressively increases with hierarchy levels, tentatively defined as twice that of the preceding level. For instance, when $m=2$, the learning rate for prototypes at the higher level is set to 0.1. The SGD optimizer is configured with momentum 0.9 to accelerate convergence through gradient averaging and weight decay 0.0001 to regularize parameter magnitudes. The proposed method is built with PyTorch on a NVIDIA RTX 3090 GPU \footnote{\url{https://github.com/HaobingLiu/HCAL}}.
%github链接

\subsection{Performance Comparison}
Table \ref{table:compare} presents a comprehensive comparison between our proposed method HCAL and six state-of-the-art multi-task learning baselines across three datasets. Notably, CIFAR-100 and Classroom adopt a two-level hierarchical label structure, while FGVC-Aircraft utilizes the last three levels of its hierarchical structure. To ensure the comparability of experimental results, all subsequent experiments are conducted using identical random seeds.

\subsubsection{Baseline} We detail the following common multi-task models to compare
with our model as below:

{\bf MMoE} ~\cite{MMOE-KDD2018} employs a multi-gate architecture where shared expert networks extract common features, and task-specific gating mechanisms dynamically combine these experts through learned attention scores.

{\bf DMT} ~\cite{DMT-CIKM2020} utilizes a modular architecture with adaptive routing to activate task-specific subnetworks. A distillation regularization term transfers knowledge from these modules to a shared backbone, balancing specialization and generalization.

{\bf CAGrad} ~\cite{ConflictAverse-nips2021a}  resolves gradient conflicts by optimizing a worst-case perturbation of task gradients within a trust region. This conflict-averse update direction improves Pareto efficiency without modifying network architecture.

{\bf IMTL} ~\cite{IMTL-ICLR2021} automatically adjusts task weights using gradient magnitude similarity and geometrically aligns gradient directions through a cosine similarity metric to reduce optimization conflicts.

{\bf Nash-MTL} ~\cite{Nash-ICML2022} formulates multi-task learning as a bargaining game, where tasks negotiate gradient updates via Nash equilibrium. This game-theoretic approach achieves Pareto optimality with provable convergence guarantees.

{\bf CMT} ~\cite{CMT-AAAI2024} proposes a task-relatedness guided multi-task learning framework that enables cross-task knowledge transfer through distribution matching loss and soft co-annotation loss, effectively solving negative transfer in scenarios with non-overlapping or partially-overlapping annotations.

\subsubsection{Main Result}
As evidenced in Table \ref{table:compare}, HCAL demonstrates comprehensive superiority across all evaluation metrics. 

On CIFAR-100, the proposed approach HCAL achieves 0.7480 $ACC^1$ and 0.7832 $ACC^2$, surpassing all baselines by a significant margin. It represents 1.40\% $ACC^1$ relative performance gains over the strongest baseline CMT. Besides, our $HVR=0.0582$ also attains an absolute reduction compared to the baselines, indicating substantially improved semantic consistency.

For FGVC-Aircraft, we achieves state-of-the-art performance with $ACC^2=0.6829$, $ACC^3=0.8233$ and $ACC^4=0.8806$, while reducing $HVR$ to 0.1012, a 9.3\% improvement over the previous best result CAGrad. Notably, it improves $ACC^2$ by 4.68\% over IMTL, confirming its efficacy in handling domain-specific feature hierarchies under limited data. This performance margin highlights our method's exceptional capability in fine-grained recognition tasks requiring precise hierarchical alignment as well.

The Classroom dataset evaluation further validates practical efficacy. HCAL maintains consistency on the Classroom dataset, attaining peak $ACC^1=0.9491$ and $ACC^2=0.9807$ with the lowest $HVR=0.0114$. The consistent $HVR$ reductions across all benchmarks confirm our approach's robustness in maintaining consistency without compromising classification precision. Although all methods perform well, the 0.16\% $ACC^2$ gain of HCAL over CMT and the minimized $HVR$ highlight its stability in near-saturated scenarios.

These results collectively demonstrate that HCAL effectively resolves the accuracy-violation trade-off in HMC, particularly on more complex and challenging datasets. Additionally, the superior performance on the real-world Classroom dataset underscores the method's practical value in authentic application scenarios.

\begin{figure*}
    \centering
    \subcaptionbox{The influence of hyperparameter $\gamma$ when $\epsilon=0.1$.\label{fig:hyper1}}{
        \includegraphics[width=0.4\linewidth]{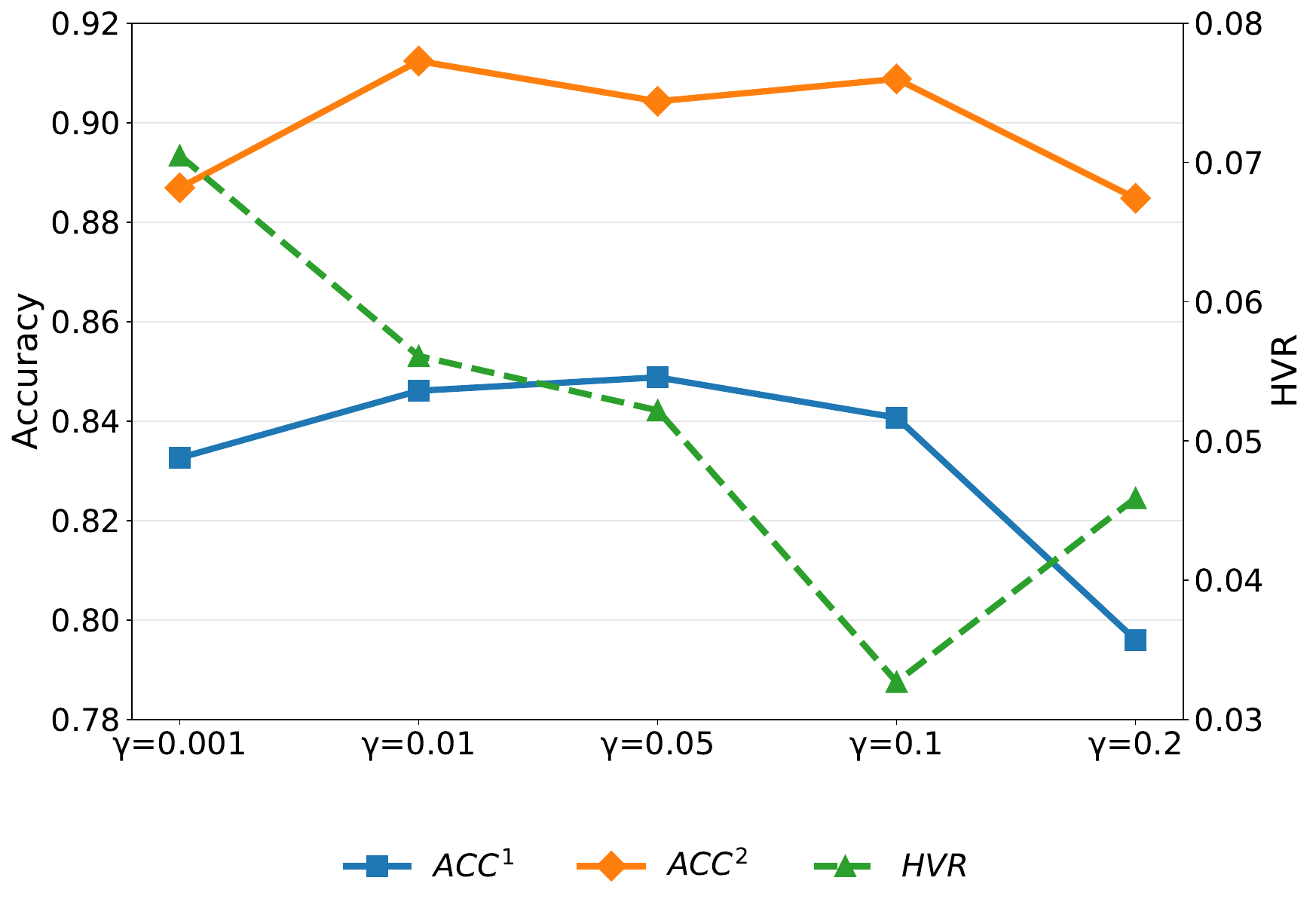}
    }
    \hspace{1cm}
    \subcaptionbox{The influence of hyperparameter $\epsilon$ when $\gamma=0.5$.\label{fig:hyper2}}{
        \includegraphics[width=0.4\linewidth]{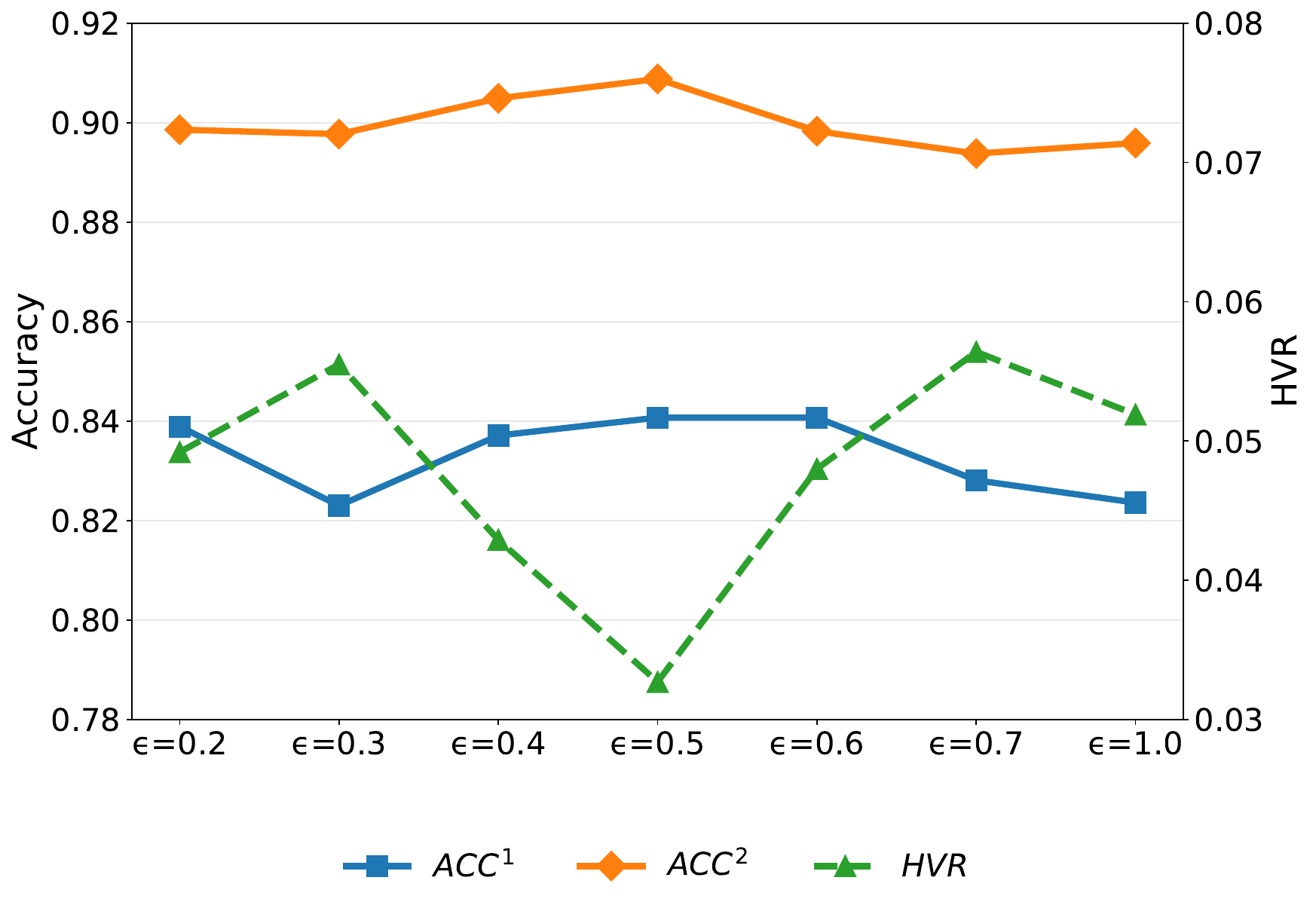}
    }
    \caption{Hyperparameter experiment on FGVC-Aircraft dataset.}
    \label{fig:hyper}
\end{figure*}

\subsection{Ablation Study}%\label{sub3}
The key components of our approach include feature aggregation, prototype perturbation, adaptive weight adjustment mechanism and multi-task learning framework. The results of the ablation experiment are reported in the level 3 and level 4 of the FGVC-Aircraft data set to verify the effectiveness of these modules. 

\subsubsection{Multi-Task Learning}
To validate the necessity of our multi-task learning framework, we conduct an ablation study by replacing it with a single-task learning framework. Concurrently, the Adaptive Weight Adjustment module is consequently rendered inapplicable in this configuration because it is intrinsically dependent on the multi-task learning framework.
Specifically, we decompose the hierarchical labels of the dataset into isolated classification tasks. Each single-task model independently predicts labels for one hierarchical level, after which $ACC$ and $HVR$ metrics are computed to quantify performance degradation.

Compared to the multi-task learning framework without the Adaptive Weight Adjustment module, the single-task learning framework exhibits a 49.2\% decline in $ACC^3$ and a 40.4\% reduction in $ACC^4$. Most critically, the $HVR$ demonstrates a staggering 19.7-fold. These findings conclusively validate that the multi-task learning framework not only achieves superior predictive accuracy but also rigorously preserves hierarchical consistency.
\subsubsection{Feature Aggregation}
This module generates hierarchical representations through multi-granularity feature fusion. In the ablation study, we remove the average pooling operation and directly utilize the backbone's output features for coarse-grained contrastive learning. Specifically, instead of aggregating features by class-wise averaging, we compute instance-level similarities between raw features and coarse prototypes. 
Disabling this operation in row 2 degrades $ACC^3$ by 2.37\% while increasing $HVR$ by 119.8\%. This demonstrates its critical role in stabilizing feature through different levels prototypes, violating parent-child relationships.
\subsubsection{Prototype Perturbation}
To evaluate the perturbation mechanism, we disable the uniform noise injection ($\epsilon$=0) and employ static prototypes for contrastive learning. Static prototypes without perturbation in row 3) reduce $ACC^4$ by 3.64\% and exacerbate $HVR$ by 129.0\%, indicating that controlled noise augmentation improves the robustness of the model in the feature space.
\subsubsection{Adaptive Weight Adjustment}
We investigate the fixed-weight baseline by replacing the dynamic adjustment mechanism with constant weighting factors, which means we let $\lambda_1=\lambda_2=0.5$. Fixed-weight baselines in row 5 increase $HVR$ by 101.8\% despite comparable $ACC^3$, revealing the necessity of adaptive balancing between levels learning objectives. Our proposed adjustment module automatically tunes the weight based on prototype gradient magnitudes, effectively resolving the conflict between hierarchical learning tasks.

\begin{table}[H]
    \centering
    \caption{Ablation experiments on the FGVC-Aircraft dataset, include the module Feature Aggregation (FA), Prototype Perturbation (PP) and Adaptive Weight Adjustment (AWA). In addition, MTL represents Multi-Task Learning. }
    \begin{tabular}{ccccccc}
    \toprule
     {\bf MTL} & {\bf FA} & {\bf PP} & {\bf AWA} & {\bf$ACC^3$ ↑} & {\bf$ACC^4$ ↑} & {\bf$HVR$ ↓} \\
    \midrule
    \XSolidBrush & \CheckmarkBold & \CheckmarkBold & \XSolidBrush & 0.4227 & 0.5331 & 0.9922  \\
    \CheckmarkBold & \CheckmarkBold & \XSolidBrush & \XSolidBrush & 0.8065 & 0.8785 & 0.0707  \\ 
    \CheckmarkBold & \XSolidBrush & \CheckmarkBold & \XSolidBrush & 0.8170 & 0.8821 & 0.0718  \\ 
    \CheckmarkBold & \XSolidBrush & \XSolidBrush & \CheckmarkBold & 0.8002 & 0.8768 & 0.0748 \\ 
    \CheckmarkBold & \XSolidBrush & \CheckmarkBold & \CheckmarkBold & 0.8218 & 0.8956 & 0.0675  \\
    \CheckmarkBold & \CheckmarkBold & \XSolidBrush & \CheckmarkBold & 0.8209 & 0.8838 & 0.0660 \\
    \CheckmarkBold & \CheckmarkBold & \CheckmarkBold & \XSolidBrush & 0.8313 & 0.8952 & 0.0502 \\
    \CheckmarkBold & \CheckmarkBold & \CheckmarkBold & \CheckmarkBold & {\bf0.8407} & {\bf0.9088} & {\bf 0.0327} \\
    \bottomrule
    \end{tabular}
    \label{tab:aba}
\end{table}

\begin{figure}[ht]
\centering
\subcaptionbox{The initial lower-level features and prototypes.\label{fig:vision2a}}{
  \includegraphics[width=0.47\linewidth]{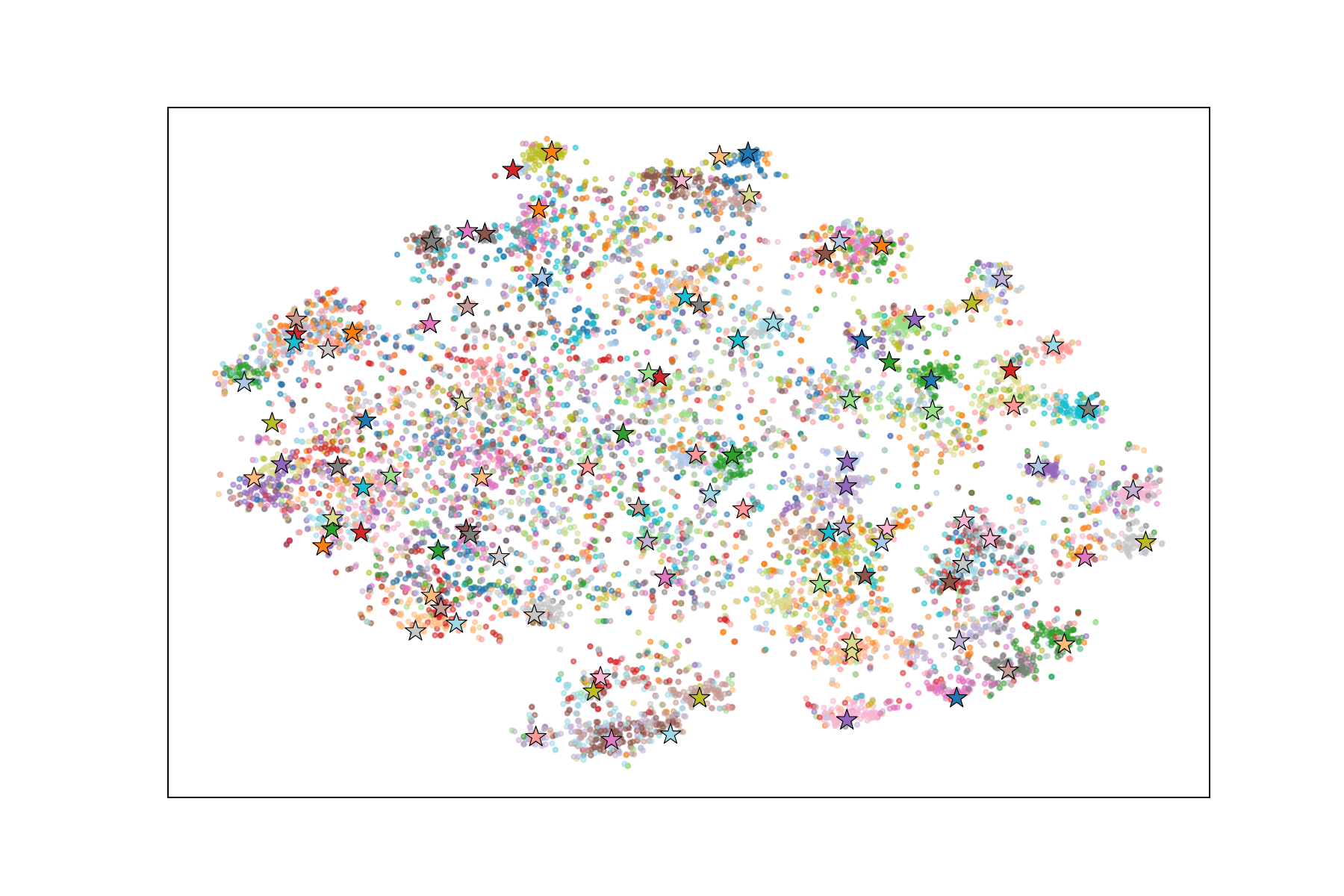}
}
\hfill
%\hspace{0.01\linewidth}
\subcaptionbox{The final lower-level features and prototypes.\label{fig:vision2b}}{
  \includegraphics[width=0.47\linewidth]{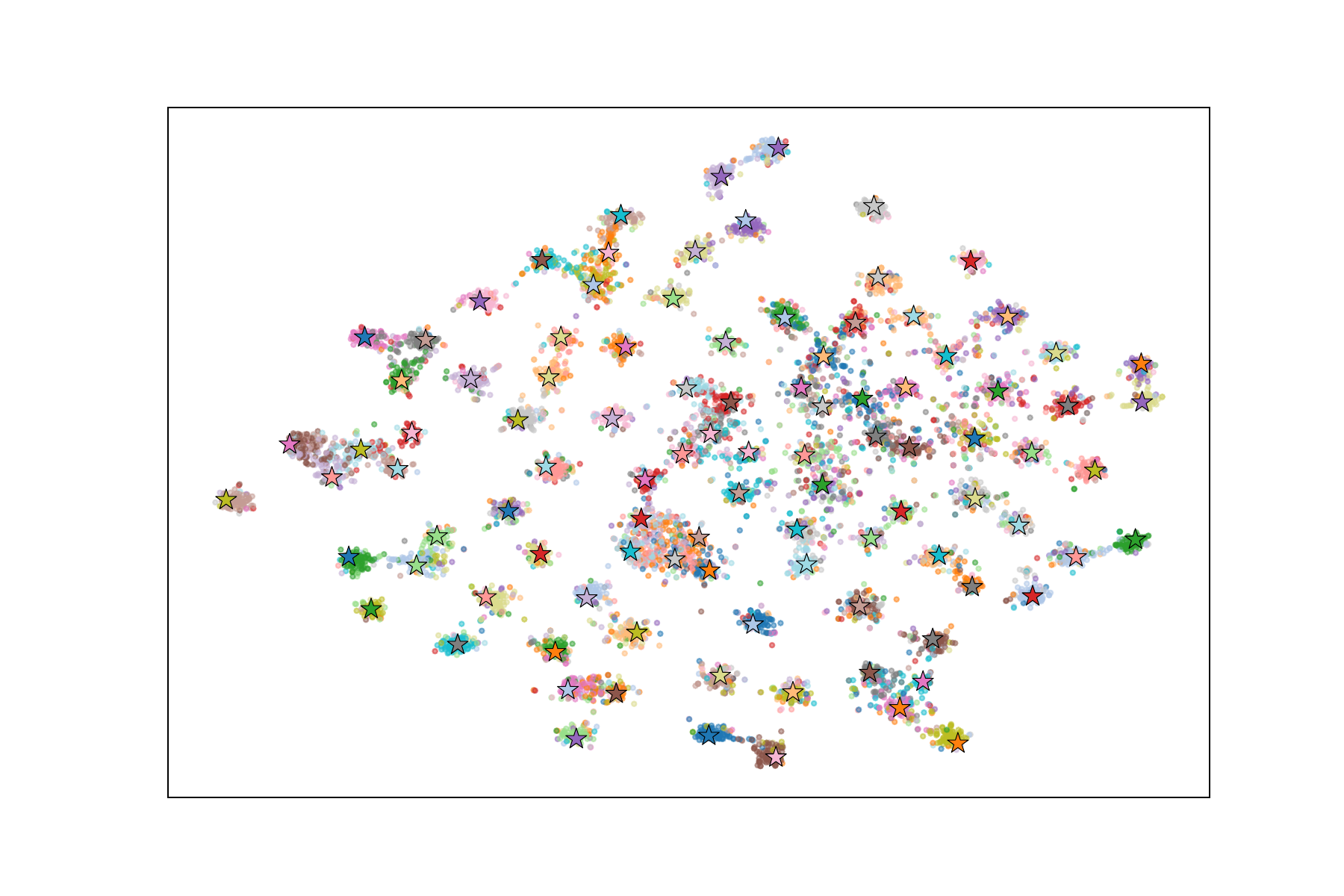}
}

\subcaptionbox{The initial higher-level features and prototypes.\label{fig:vision2c}}{
  \includegraphics[width=0.47\linewidth]{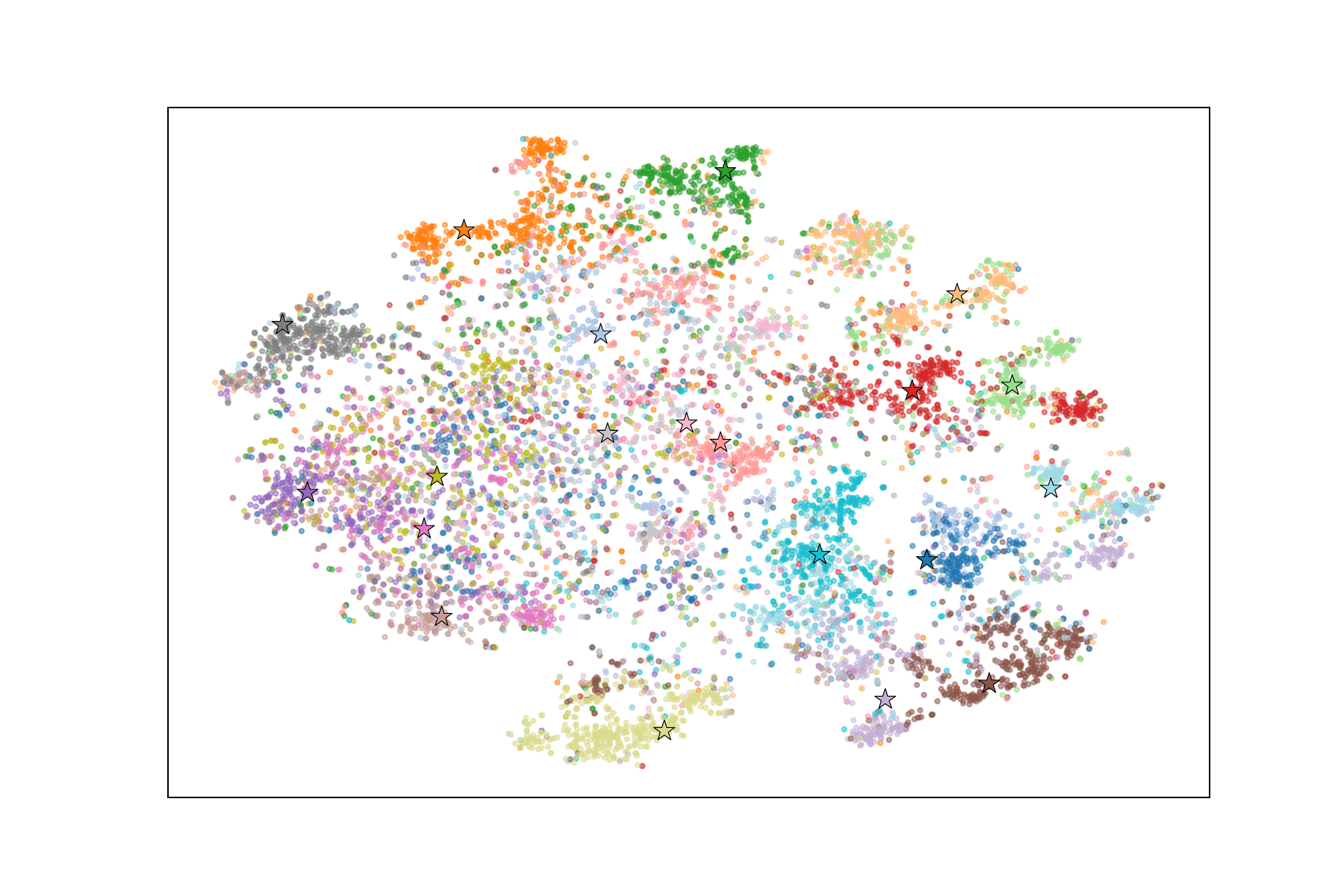}
}
\hfill
%\hspace{0.01\linewidth}
\subcaptionbox{The final higher-level features and prototypes.\label{fig:vision2d}}{
  \includegraphics[width=0.47\linewidth]{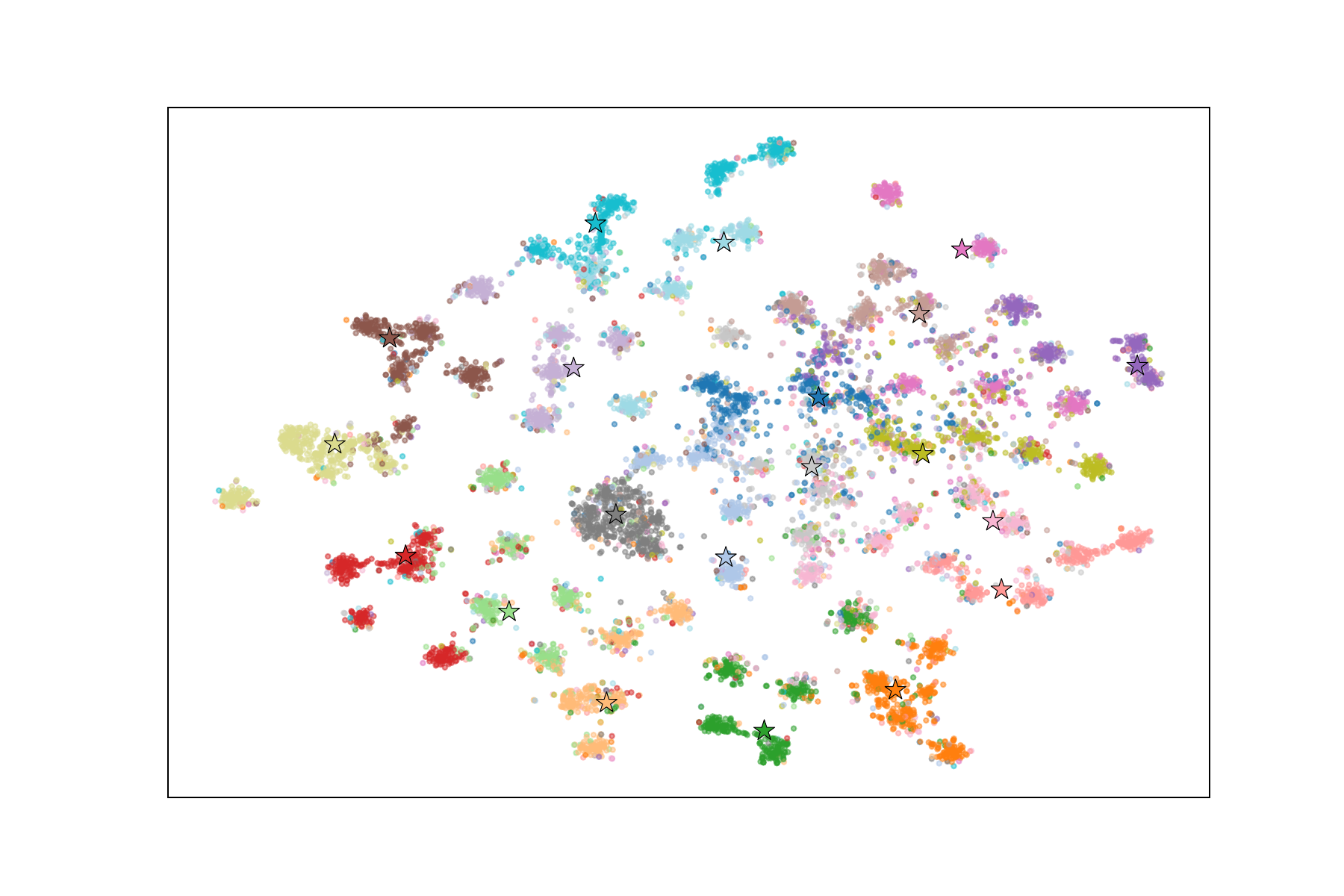}
}

\caption{Features and prototypes visualization of CIFAR-100 dataset, where color-coded points denote features from distinct categories and colored five-pointed stars represent prototypes.}
\label{fig:vision-2}
\end{figure}

\begin{figure}[htbp]
\centering
\subcaptionbox{The initial features and prototypes of two levels.\label{fig:vision3a}}{
  \includegraphics[width=1\linewidth]{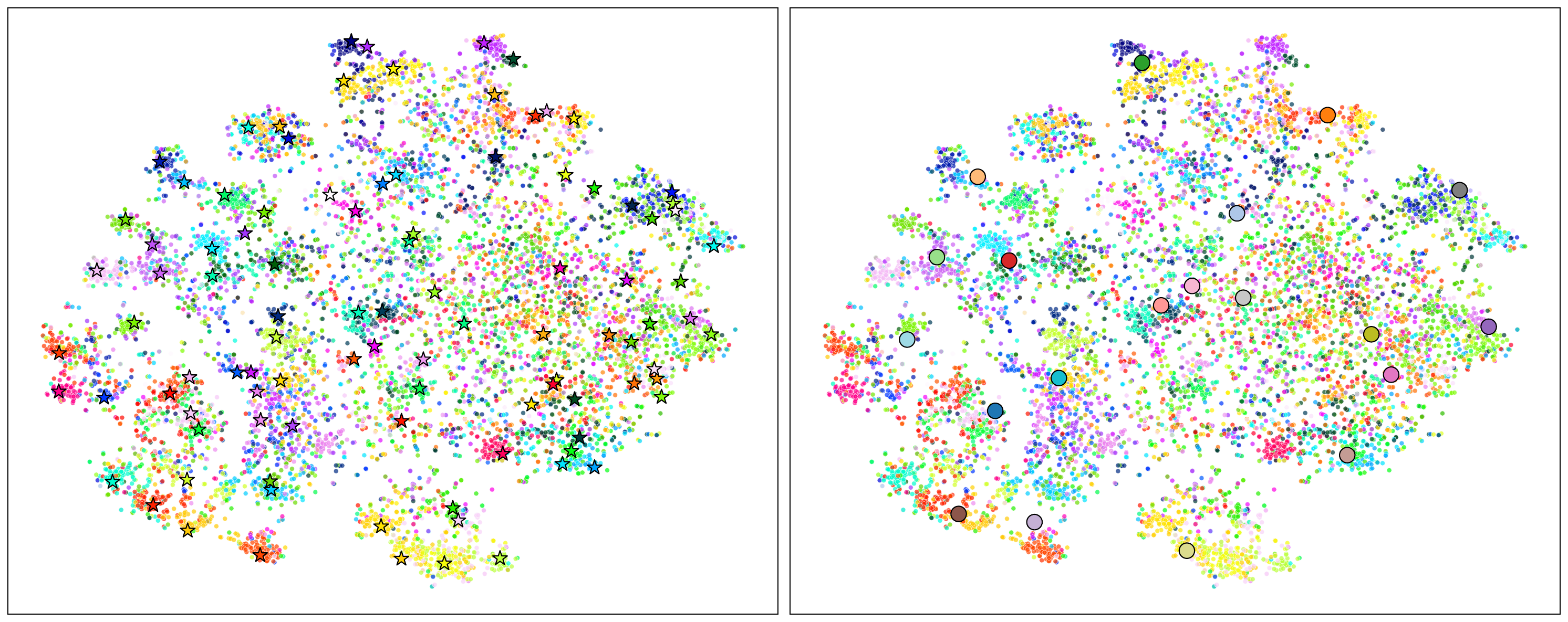}
}
\subcaptionbox{The final features and prototypes of two levels.\label{fig:vision3b}}{
  \includegraphics[width=1\linewidth]{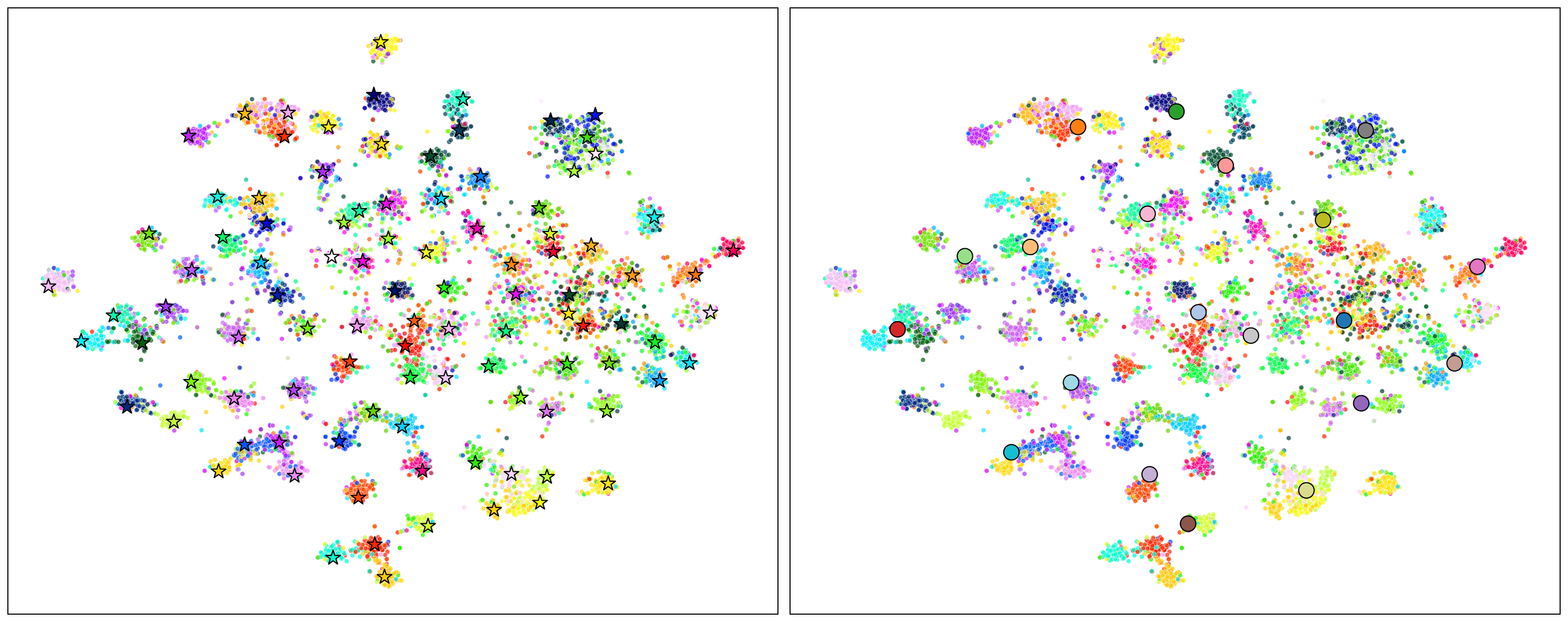}
}
\caption{Features and prototypes visualization of CIFAR-100 dataset, where the features are same in subgraph. The five-pointed stars represent lower-level prototypes and circles denote higher-level prototypes.}
\label{fig:vision-3}
\end{figure}

\subsection{Hyperparameter Analysis}
Hyperparameters are non-trainable parameters in machine learning models that must be manually specified prior to the training phase. In our experiments, the hyperparameters $\epsilon$ in prototype perturbation and $\gamma$ in adaptive weight adjustment mechanism have obviously affected the results. Similar as ablation study, we leverage the level 3 and level 4 of FGVC-Aircraft to report our results. The influence is shown in Figure \ref{fig:hyper}.

Figure \ref{fig:hyper1} shows the impact of hyperparameter $\gamma$ on $ACC^1$, $ACC^2$ and $HVR$ on the FGVC-Aircraft dataset. The $\gamma$ serves as a critical hyperparameter governing multi-task weight allocation. By modulating the softmax distribution of task weights through temperature scaling, it dynamically adjusts the sharpness of weight assignments, thereby regulating training stability. Experimental results reveal $\gamma=0.5$ achieves optimal balance with $ACC^1=0.8407$ and $HVR=0.0327$. A higher $\gamma$ value flattens the weight distribution to promote balanced task learning, while a lower $\gamma$ sharpens the focus on dominant tasks, mitigating gradient conflicts. 

Figure \ref{fig:hyper2} analyzes the hyperparameter sensitivity of prototype perturbation intensity $\epsilon$ on the FGVC-Aircraft dataset. As the governing parameter for noise injection, $\epsilon$ nonlinearly modulates model robustness. Moderate values such as 0.01 to 0.1 balance discrimination and generalization through controlled spherical perturbations, achieving peak performance at $\epsilon=0.05$ with $ACC^1=0.8488$. Excessive noise levels like $\epsilon=0.2$ degrade semantic coherence, reducing $ACC^1$ by 4.88\% and $ACC^2$ by 2.76\%. Conversely, insufficient perturbation with $\epsilon=0.001$ yields suboptimal discrimination at $ACC^1=0.8326$. These results confirm that perturbation prevents overfitting while preserving categorical structures via contrastive augmentation.

\subsection{Visualization}%\label{sub5}
To visually assess the hierarchical classification efficacy of our approach, we conducted comprehensive visualizations on both feature embeddings and prototype distributions for CIFAR-100 under a two-level hierarchy. The results are systematically presented in Figure \ref{fig:vision-2} and Figure \ref{fig:vision-3}.

As illustrated in Figure \ref{fig:vision2a} and Figure \ref{fig:vision2c}, during the early training phase, both feature embeddings and their corresponding prototype positions of two levels demonstrate substantial overlapping distributions. This indicates ambiguous decision boundaries and inefficient task-specific discriminability before optimization. Conversely, in Figure \ref{fig:vision2b} and Figure \ref{fig:vision2d}, we can find that features cluster tightly around their corresponding prototypes with clear separation between subcategories, which evidences usefulness of training.

Furthermore, Figure \ref{fig:vision-3} provides critical insights into hierarchical consistency, where we guarantee that the color and position of the features are exactly the same in the left and right images. It is obvious to find higher-level prototypes initially lack structural alignment with their child classes in Figure \ref{fig:vision3a}. After full optimization, exhibited in Figure \ref{fig:vision3b}, parent-class prototypes converge to the geometric centroids of child-class clusters, achieving structural congruence between hierarchical levels. 

These visualizations corroborate that our approach achieves high classification accuracy and adherence to hierarchical constraints.

\section{Conclusion}
In this paper, we propose the classifier called HCAL based on MTL and prototype contrastive learning with adaptive loss weight adjustment mechanism. It generates multi-level feature representations from inputs through the backbone network and initializes category-specific prototype vectors with perturbation. It also optimizes features and prototypes jointly by using a hierarchical contrastive loss and gains the output maintaining hierarchical consistency. Extensive experiments have demonstrated that results from the our method achieve a better performance than existing methods in both classification accuracy and hierarchical constraint adherence.

\begin{acks}
This research is supported in part by the National Science Foundation of China (No. 62302469, No. 62176243), the Natural Science Foundation of Shandong Province (ZR2022QF050, ZR2023QF100), and the Fundamental Research Funds for the Central Universities (202513026).
\end{acks}

\bibliographystyle{ACM-Reference-Format}
\bibliography{cite}

\end{document}